# Snail Homing and Mating Search Algorithm for Weight Optimization of Stepped-Transmission Shaft

Kaustav Saha[1], Ishaan R Kale[2]*, Vivek Patel[1], Anand J Kulkarni[2], Puskaraj D Sonawwanay[1]

[1]Department of Mechanical Engineering, Dr Vishwanath Karad MIT World Peace University, Kothrud, Pune, India

[2]Institute of Artificial Intelligence, Dr Vishwanath Karad MIT World Peace University, Kothrud, Pune, India

*Corresponding Author: ishaan.kale@mitwpu.edu.in*

**Abstract**

In this paper, the steeped-transmission shaft design problem is proposed for weight optimization. The bio-inspired search-based Snail Homing and Mating Search (SHMS) algorithm is utilized to solve the problem. It is inspired by the social behaviour of snails and their inherent nature of finding better homes, and mate. The proposed steeped-transmission shaft design problem is modelled considering the fatigue loading, combined bending, torsion loads, and the principle of Modified Goodman criteria. The forces diagram and the bending moment diagrams are obtained using the MDSOLIDS software. The forces and bending moment are then used to mathematical model the objective function and constraints. The SHMS algorithm has yielded the desired solution with reasonable computational cost. The constraints are handled using a static penalty function approach. The statistical results obtained using SHMS algorithm are further used for generating CAD model. The analysis is carried out in ANSYS Workbench. Further, the deflection obtained from SHMS algorithm and ANSYS Workbench are compared and results are discussed in details.

## 1. Introduction

Mechanical components are designed to withstand under static/dynamic loading condition. These loading conditions are often occurred in combination which significantly affect the performance, strength, and various other mechanical aspects of the component (Norton, 2010). The precise design of the mechanical components helps to sustain in diversified conditions. For that optimization plays a crucial role which leads to cost reduction by minimizing the material usage, streamlining manufacturing process and decreasing the need for frequent maintenance. This not only saves resources, but also contributes to a more sustainable approach to engineering. In real-world design problems, numerous parameters influence the objective function in a highly intricate and nonlinear manner. Furthermore, optimized components are inherently safer due to their improved reliability and reduced risk of failure. The processes of optimization include usage of Finite Element Analysis (FEA), or usage of optimization algorithms to obtain desired numerical solution. The important aspect of the mechanical component is to optimize (minimize) the weight. Weight reduction has numerous benefits such as increased efficiency due to low power consumption and less inertia to accelerate, lower material costs due to low material usage, longer lifespan and improved aesthetics. In the aerospace industry, components like wings and the aircraft body must be as light as possible to save fuel and increase the amount of cargo they can carry. This leads to lower operating costs and a smaller environmental impact. Similarly, in the automotive industry, lighter engine parts, suspension systems, and body panels improve fuel efficiency, acceleration, and handling. Industries that apply robotics and renewable energy also benefit from lightweight designs (Zhu et al., 2021; Yao et al., 2019). Robotic arms, wind turbine blades, and solar panel structures can be made lighter to improve their mobility, efficiency, and reduce manufacturing costs. Weight reduction of transmissions and drive shafts is a crucial factor in improving vehicle or power transmission system performance and efficiency

Analytical and numerical methods are often effective for finding extreme values of a function in practical applications however, they may struggle while solving more complex design scenarios. In Mechanical components, weight reduction is generally preferred criteria for optimization. In such cases, advanced optimization algorithms offer viable solutions by efficiently identifying near-optimal results within reasonable computational time and cost (Rao and Savsani, 2012). Metaheuristic algorithms inspired form natural phenomenon such as Genetic Algorithm (GA) (Deb and Goel, 1996; Guzmán and Delgado, 2005), Particle Swarm Optimization (PSO) (Rodriguez-Cabal, 2018), Probability Collectives (Kulkarni et al., 2013, 2016), Continuous Genetic Algorithm (CGA) (Cabal et al., 2019), Cohort Intelligence (CI) Algorithm (Kale and Kulkarni, 2018, 2021a, 2021b), Whale Optimization Algorithm (WOA) (Mirjalili and Lewis, 2016; Ning and Cao, 2021), Vortex Search Algorithm (VSA) (Rodriguez-Cabal et al., 2021), Teaching Learning Based Optimization (TLBO) (Rao et al., 2011), Soccer League Championship Algorithm (Moosavian and Roodsari, 2014), Symbiotic organisms search (SOS) (Cheng and Prayogo, 2014), Modified Leader-Advocate-Believers (LAB) Algorithm (Reddy et al., 2024) are well examined for solving the problems from mechanical engineering domain.

In this paper, a new mathematical model of the power transmission stepped shaft is presented designed with varying diameters along their length, which allows for increased strength in areas of high stress. The stepped design provides effective axial load support to the power transmission process. Increasing the diameter in critical sections, the shaft's resistance to bending and torsional moment is significantly enhanced. This prevents the shaft from deforming or failing under the applied forces (Jweeg et al., 2020). Due to various types of rotating components and combined loading (bending and twisting), the stepped shaft is able to counter stress concentrations in different positions throughout its length. The steeped shaft shoulders created by the diameter changes act as robust support points for components like gears and pulleys. These shoulders prevent axial movement of the components, ensuring that they remain securely positioned and aligned on the shaft. This is particularly important in applications involving helical or bevel gears, which generate significant axial forces during operation. Transmission shafts undergo mechanical defects such as deflection, vibration, noise, wearing, fatigue, etc. Minimizing deflection is one of the most important criteria for mechanical component optimization as it reduces misalignment of gears, pulleys, or other components mounted on the shaft. Deflection can cause components to contact inadequately, leading to friction and wear, as well as potential damage to the components (Zhang et al., 2022). The proposed power transmission stepped shaft problem is solved using a bio-inspired Snail Homing and Mating Search (SHMS) algorithm (Kulkarni et al., 2024). The constrained along with the objective function are handled using the Static Penalty Function approach (SPF). Whenever the constraint equations violate the relation, a penalty value is imposed on the objective function. This help in navigate the solution towards feasibility. The results obtained from SHMS algorithm is validated using the Finite Element Analysis (FEA) ANSYS Software.

Throughout the discussion, formulation of mechanical component, mathematical model and the solution of the selected mechanical component using SHMS algorithm has been discussed. The mechanical component selected for the study is explored as mechanical design problem. A comprehensive literature review covering previous studies on metaheuristic algorithms for the optimization of mechanical components is presented in Section 1.1. The application of stepped transmission shafts in mechanical power transfer has been thoroughly discussed in Section 1.2. This section also provides a detailed explanation of how the bending moments and torque are evaluated to understand their effects on the shaft, and how to determine safe diameters using Von Mises stresses in conjunction with the Modified Goodman Criterion. In Section 2, the SHMS algorithm is introduced, along with a brief description of its biological inspiration from the behaviour of snails, particularly their continuous processes of homing and mating. The robust working principle of the algorithm is outlined in Section 3, and its detailed mathematical formulation is discussed in Section 3.1. The problem of stepped transmission shaft and its initial setup is presented in Section 4. The mathematical model of the problem statement containing all the loads, boundary condition and physical quantities is introduced in section 4.1 with the problem statement formulated using the SHMS algorithm through the Static Penalty Function (SPF) approach, as detailed in Section 4.2. In Section 5, the problem is addressed using both analytical and simulation (FEA) methods. Section 5.1 explains the analytical approach, including the free body diagram of the rotating elements (gear and pulley) by identifying the forces acting on the shaft during rotation. The calculated forces acting on the stepped shaft are then used in MDSOLIDS to determine the bending moments. These bending moment values are subsequently used in the SHMS algorithm to evaluate the constraints. The shaft diameters optimized using the SHMS algorithm are then used for modelling in ANSYS Workbench, where simulations are conducted to analyse the rotating motion under an applied torque, as covered in Section 5.2. The resulting deflection values from ANSYS are compared in detail with those obtained from the SHMS algorithm in Section 6. This section also provides an in-depth analysis of three cases; 'Best,' 'Mean,' and 'Worst'—based on maximum power transmission, comparing both analytical results and visual outputs from ANSYS Workbench. The deflection values obtained in each case from ANSYS are analysed, and a suitable conclusion is drawn at the end of the discussion regarding the optimization of the stepped transmission shaft in Section 7.

## 1.1 Literature Review

In the recent studies, the metaheuristic algorithms are applied for several mechanical engineering problems for minimization of weight, deflection and cost. The nature inspired metaheuristic algorithms are designed to solve complex problems that traditional methods cannot effectively solve (Plevris and Solorzano, 2022). The Modified Bacterial Foraging Optimization (BFO) algorithm is a metaheuristic algorithm that handles constrained problems using SPF approach in various engineering design problems (Mezura-Montes and Hernández-Ocaña, 2009). Symbiotic Organisms Search (SOS) is another metaheuristic algorithm which was used to solve 26 multi-dimensional unconstrained optimization problems and achieved appreciable results compared to Particle Swarm Optimization (PSO) (He and Wang, 2007), GA, Bees Algorithm (BA), Differential Evolution (DE) and Particle Bee Algorithm (PBA) and was later used to solve stepped cantilever beam, I-section beam and multi-bar truss problems (Cheng and Prayogo, 2014). Firefly Algorithm (Gandomi et al., 2011), CI Algorithm (Kale and Kulkarni, 2018) for handling discrete and continuous variables for linear and non-linear constraints using SPF approach provided solid background to proceed with the similar optimization of the mechanical component, a stepped shaft. These algorithms are inspired by natural phenomena such as evolution, genetics, and swarm behaviour, and social behaviour of humans, enabling them to explore vast search spaces to find the global optimum of a given problem. These algorithms have very diverse application and are not constrained to a particular case. Due to their diverse application, they have slowly found their way in engineering and technology since the last decades for the purpose of optimization. In case of power transmissions, Stepped Transmission shafts are preferred as a mechanical component due to its simplicity and improved stress distribution. In Guzmán and Delgado (2005), the first problem statement was devised for optimization of the stepped shaft made of AISI 1040 CD Steel, using GA multi-objective optimization. The results gave us the various diameters of the stepped shafts that produce minimal deflection upon loading and power transmission. Other studies show a well-known metaheuristic algorithm known as PSO has been used for optimization of the similar shaft with same loads and boundary conditions without consideration of deflection (Rodriguez-Cabal, 2018). This algorithm gave improved and more optimised values of diameters which resulted in lesser weight. Another very recent approaches to optimization of such stepped shafts, called CGA (Cabal et al., 2019) and VSA have been implemented (Rodriguez-Cabal et al., 2021). It considered the deflection of the shaft permissible within 0.005 inches. This caused the algorithm to become more integrated and gave not only lesser weight of the stepped shaft, but also deflections within the permissible limit with same loads and boundary conditions. Furthermore, the stepped shaft material was varied and composite materials were used in place of conventional steel material with slight change in the shaft geometry, and a new algorithm was introduced known as Ant Lion Optimization (ALO) (Rodriguez-Cabal et al., 2024). At first ALO was used in the discussion and was compared with CGA, PSO and VSA in great detail on the basis weights of the shafts. Later, the Carbon Fiber Reinforced Polymer (CFRP) material properties were used to design the shaft with the help of the above discussed algorithms and it evaluated appreciable results with lesser weights compared to the results obtained with AISI 1040 CD Steel. Additive manufacturing allows for intricate designs and tailored material distributions, while CFRP composites offer superior strength-to-weight ratios, making them ideal for lightweight, high-performance applications. However, their use in power transmission systems presents challenges due to anisotropic properties, manufacturing limitations, and complex material behaviour.

## 1.2 Background of Study

The mechanical component discussed throughout the context is a transmission shaft, also referred to as the drive shaft which transmit power from one component to another component in a mechanical system by transferring the rotational motion between different elements such as pulleys, gears and couplings. It is usually a cylindrical rod of uniform diameter with high tensile and compressive strength which is able to withstand high amounts of bending, torsion, shear, tension, etc. In some cases, it is found that there are more than one type of forces acting on the shaft, which may subject it to fluctuating loads. In such cases, stepped shafts are designed. A stepped transmission shaft consists of multiple sections with varying diameters along its length. This design is essential for fitting different machine components like bearings, gears, and pulleys, which often require specific mounting dimensions. The stepped structure helps minimize stress concentrations, ensuring better load distribution while maintaining strength and reducing overall weight (Rodriguez-Cabal et al., 2024). Furthermore, it enhances alignment and stability, particularly in high-speed applications, effectively reducing vibrations and improving the mechanical system's overall performance.

To calculate the effective diameters of the shaft, the principle of combined loading on a shaft using Modified Goodman criteria for fluctuating load is considered (refer Fig 1). Since the power fluctuates between 5 hp and 20 hp, there is a fluctuating twisting moment as well as bending moment, referred to as $T$ and $M$ (Guzmán and Delgado, 2005). The alternating and mean components of such twisting moment is denoted as $T_a$ and $T_m$ respectively, whereas the alternating and mean components of bending moment is denoted as $M_a$ and $M_m$ respectively. In Mechanical engineering applications, the Goodman line is applied to predict the fatigue life of components exposed to cyclic loading conditions. These loads, like those encountered in rotating shafts or machinery, fluctuate repeatedly. Fatigue failure can occur even when the maximum stress in each cycle remains below the material's yield strength, resulting from the cumulative effect of these repeated stress cycles.

The Goodman line is a visual tool that helps to forecast the type of failure. It graphically illustrates the relationship between the mean stress (the average stress within a load cycle) and the alternating stress (the magnitude of stress fluctuations around the mean) for a specific material. By analysing the position of the actual stress conditions (represented as an operating point) on this diagram, engineers can assess the likelihood of fatigue failure. If the operating point falls below the Goodman line, the component is anticipated to exhibit a long fatigue life. Conversely, if it lies above the line, the risk of fatigue failure increases significantly. Whereas the Modified Goodman line is an enhancement to the original Goodman line used in fatigue analysis (Sekercioglu, 2009). It addresses the potential for yielding to occur at elevated mean stress levels, which the original Goodman line might not fully capture. This modified line is depicted as a straight line that connects two critical points: the endurance limit $S_e$ (the maximum stress amplitude a material can withstand indefinitely) on the alternating stress axis ($\sigma_a$) and the ultimate tensile strength $S_{ut}$ (the maximum stress a material can endure before fracturing) on the mean stress axis ($\sigma_m$). This adjustment provides a more accurate assessment of fatigue life, particularly in scenarios with significant mean stress components. The key reasons for its widespread application are its ability to consider mean stress effects. Unlike simpler fatigue criteria that only focus on alternating stress, the Modified Goodman diagram incorporates the detrimental influence of tensile mean stresses, which can accelerate fatigue failure. This makes it particularly useful for high-cycle fatigue applications, where stress remains within the elastic range, and the number of cycles to failure is large.

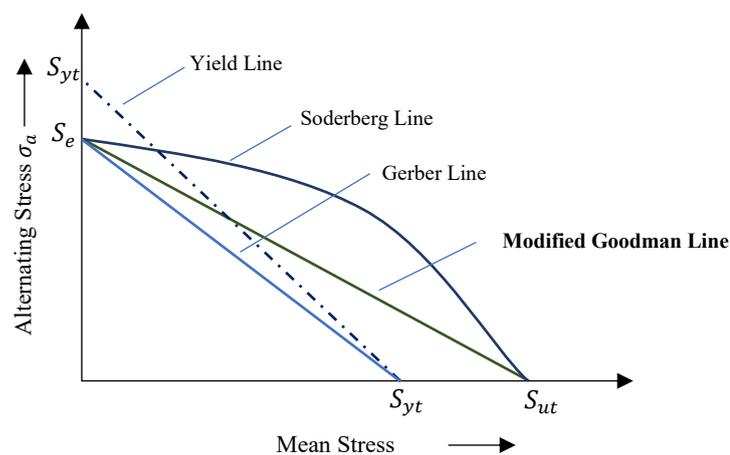

**Fig. 1** Graph of alternating stress amplitude vs mean stress amplitude which describes the Modified Goodman Line (Sekercioglu, 2009; Kumbhar and Tayade, 2014)

Modified Goodman criteria is represented in mathematical form as follows:

$$\frac{1}{N_f} = \frac{\sigma'_a}{S_e} + \frac{\sigma'_m}{S_{ut}} \qquad (1)$$

where the resultant alternating stress

$$\sigma'_a = (\sigma_a^2 + 3\tau_a^2)^{1/2} = \left[\left(\frac{32 K_f M_a}{\pi d^3}\right)^2 + 3\left(\frac{16 K_{fs} T_a}{\pi d^3}\right)^2\right]^{1/2} \qquad (2)$$

and resultant mean stress

$$\sigma'_m = (\sigma_m^2 + 3\tau_m^2)^{1/2} = \left[\left(\frac{32 K_{fm} M_m}{\pi d^3}\right)^2 + 3\left(\frac{16 K_{fsm} T_m}{\pi d^3}\right)^2\right]^{1/2} \qquad (3)$$

The alternating stress at each point varies due to variation of moment (bending and twisting) as well as different sections of shaft. The bending moment at different sections of the shaft can be achieved by the bending moment diagram (BMD) for the maximum and minimum power transfer cases. Accordingly, the mean and alternating values can be calculated as follows:

$$\sigma_a = \frac{32 K_f M_a}{\pi d^3} \qquad (4)$$

$$\sigma_m = \frac{32 K_{fm} M_m}{\pi d^3} \qquad (5)$$

Also, the shear stress developed at each section due to the alternate and mean twisting moment can be calculated as follows:

$$\tau_a = \frac{16 K_{fs} T_a}{\pi d^3} \qquad (6)$$

$$\tau_m = \frac{16 K_{fsm} T_m}{\pi d^3} \qquad (7)$$

With the obtained values of $M_a$, $M_m$, $T_a$, $T_m$, and putting the values in the modified Goodman Equation, the optimum design diameter of the stepped shaft section is calculated. The process is repeated for all the stepped sections. The expression of diameter obtained from the Modified Goodman equation is given as follows:

$$d_i = \left[\frac{32 N_f}{\pi} \cdot \left\{\frac{\sqrt{(K_f M_a)^2 + \frac{3}{4}(K_{fs} T_a)^2}}{S_e} + \frac{\sqrt{(K_{fm} M_m)^2 + \frac{3}{4}(K_{fsm} T_m)^2}}{S_{ut}}\right\}\right]^{1/3} \qquad (8)$$

where

$N_f$     Safety Factor
$K_f$     Stress concentration factor for alternating components for Bending Stress
$K_{fm}$     Stress concentration factor for mean components for Bending Stress
$K_{fs}$     Stress concentration factor for alternating components for Torsional Shear Stress
$K_{fsm}$     Stress concentration factor for shear mean components for Torsional Shear Stress

## 2. Biological Background of Snails

Snails and slugs are unique creatures with fascinating adaptations. As hermaphrodites, the snails possess both male and female reproductive organs, allowing them to reproduce independently or with a mate. These gastropods inhabit diverse environments, from aquatic bodies to terrestrial habitats. Their primary needs include food, water and shelter (Kulkarni et al., 2024). To move, they secrete a slimy substance (mucous) that aids in locomotion but also increases water loss from their bodies, making them dependent on humid conditions. Despite having eyes, their vision is limited. Instead, they rely on chemical cues and their own mucous trails to navigate their surroundings. During this movement, the snails leave behind trails of information, similar to the mucus trails left by snails. These snails are evaluated, and those deemed fecund (i.e., having high fitness) are identified from the trails they leave behind while navigation. Other snails in the vicinity are then guided towards these fecund individuals, or towards other snails based on their individual preferences. As the snail meet the fecund individuals, they initiate a mating process. During the search for its fecund pair and after mating, the snails release certain chemical signal that attracts other snails towards the successful regions of mating to happen. Following the mating process, the snails seek out for better shelters by relocating to nearby homes or locations with more favourable conditions. While searching for these better shelters, the snails encounter the trails left by other snails. These trails provide valuable guidance for the later snails, directing the search towards promising areas and the locations of successful mating events. Likewise other snails too perform the same behaviour and thus the snail's behaviour plays a very important role for constantly seeking for better and better survival and safety. This iterative process of locomotion, mating, and shelter seeking continues, with each generation of snails building upon the knowledge and experiences for the nearby snails who are seeking for better homes and habitat. This dynamic interaction and information sharing drive the snails towards increasingly better and favourable situations.

Snails choose their paths based on the presence of fresh, dense mucous trails, which help them locate new homes and follow fecund snails. Fig 2(a) illustrates how snails prefer trails laid by other snails with fresh mucous, as these provide a stronger chemical signal compared to older trails that may have been washed away by water or dispersed by wind. Snails possess highly sensitive chemoreceptors in their front tentacles, allowing them to detect chemical signals in mucous paths. As they navigate, they are more attracted to stronger scents from fresh mucous than to weaker ones and continue following the trail until they find a home better than their existing one, as shown in Fig 2(b). Snails seek habitats that are cool, moist, and shady, as these conditions helps to minimize fluid loss from their bodies. If they find the new home less suitable than their previous one, they retrace their path and return to their original home. Thus, mucous tracks not only serve as pathways to discover new homes but also act as guiding trails that help snails navigate back to their current home.

Besides homing, snails also follow scents of the mucous to track fecund snails to perform mating. While searching for new homes by moving along the trails, they may pick up fecund snail mucous trails. Snails locate fecund (highly reproductive) individuals by detecting specific chemical cues within their mucous trails. Their front tentacles are equipped with highly sensitive chemoreceptors, enabling them to recognize and differentiate between various trails. The mucous left behind by snails serves not only as a movement aid but also as a carrier of biochemical markers, such as proteins and pheromones, which indicate the reproductive condition of the snail that created the trail. When encountering multiple mucous trails, snails tend to follow those that are fresher and richer in chemical signals, disregarding older or weaker ones. Trails left by fecund snails contain reproductive pheromones that act as signals, attracting other snails and encouraging them to follow. This capability allows snails to effectively locate potential mates, even in environments where populations are sparse, by relying on these biochemical cues, as shown in Fig 2(c). The overall behaviour of snails in the ecosystem is rather a very simultaneous process, which is represented by Fig 2(d), describing Homing, Mating and trail following characteristics of snails. Every snail wants to survive in an ecosystem by finding secure homing places with perfect conditions and by mating with other snails to evolve.

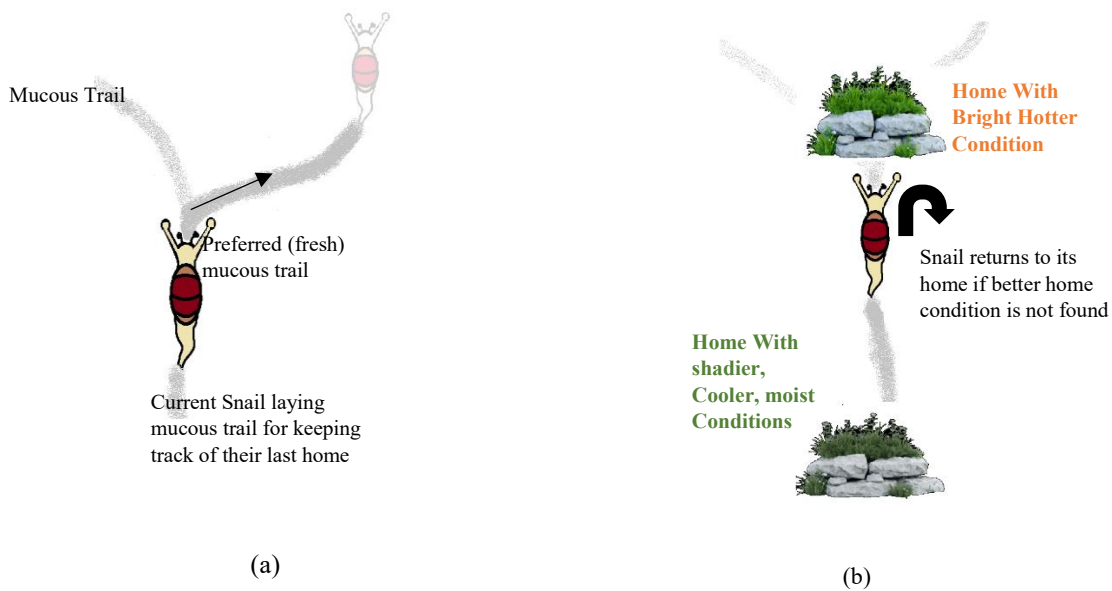

(a)  (b)

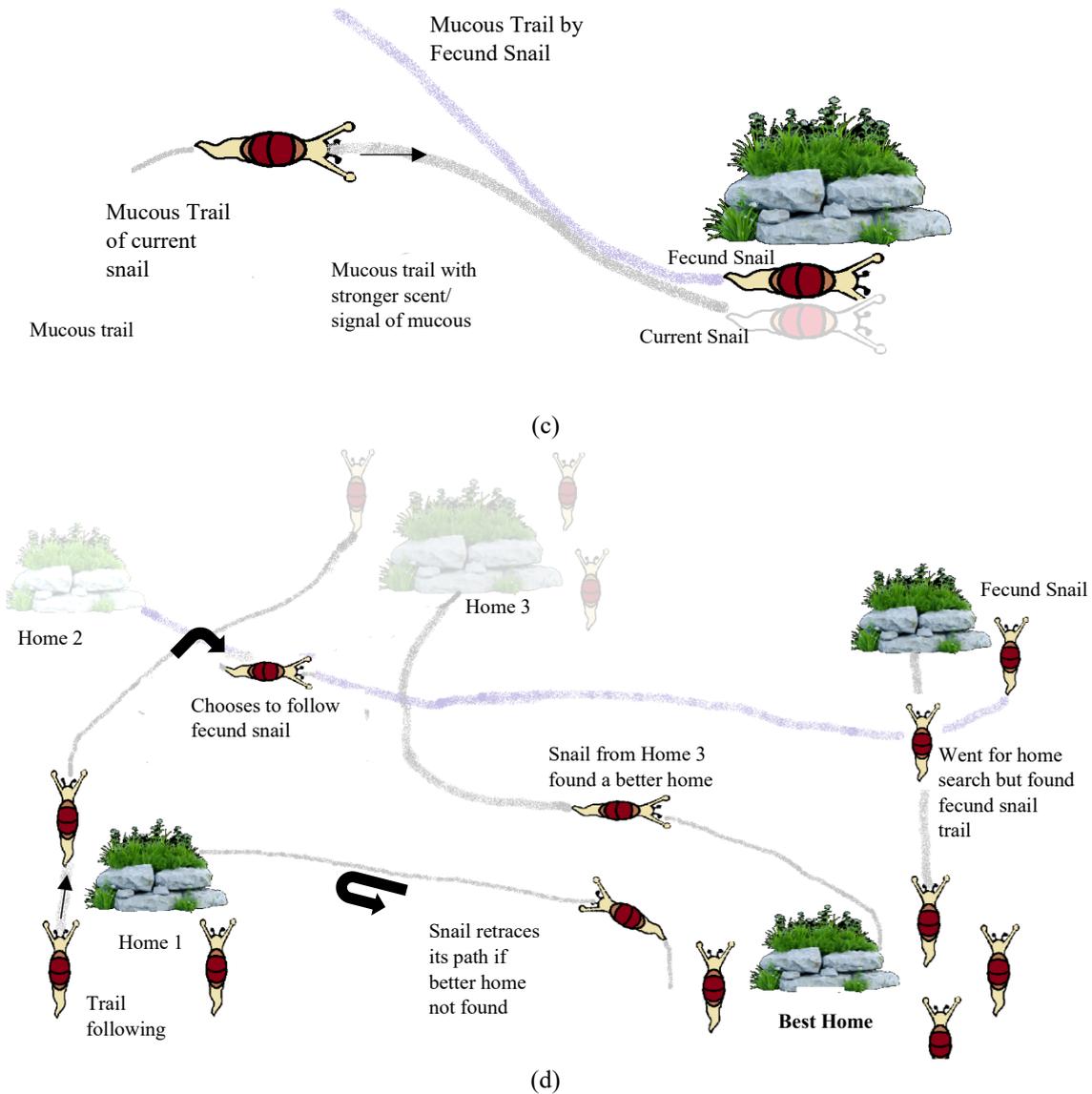

**Fig. 2** The explanation of the behaviour of snails for Homing and Mating and searching new homes
  (a) Selection of the mucous trail based on fresher mucous trails.
  (b) Selection of mucous trails based on home condition (e.g., Dark, Cool, moist, etc.)
  (c) Location of fecund snails by their mucous trails
  (d) Generation of snails around the homes and they repeat the above behaviour.

## 3. SHMS Algorithm

The SHMS algorithm is bio-inspired metaheuristic algorithm. It mimics the snail behaviour of homing, food search, mate search and trail following (Kulkarni et al., 2024). The working mechanism of the algorithm starts with generation of snails around homes. For better habitat and the mate, they follow the available trail on the ground. The snails are continuously looking for the trail with better mucus quality which laid by the fecund snail. This helps them to find the fecund snail for mating and sign for a good habitat while returning home. The snails continuously seek improved survival conditions and potential mates by following mucous trails left behind for mating. To translate this behaviour into an optimization algorithm, the concept of generating snails around their homes is introduced, utilizing a sampling interval that updates with each iteration. The fecundity of the snails is determined based on their fitness values-an increasing fecundity value indicates an improving solution, while a negative value suggests a decline in solution quality. The selection of mating pairs follows a probabilistic approach, where snails have a certain probability of being paired with their fecund partner. The placement of snails based on fecundity values is carried out using the Roulette Wheel Selection method. Additionally, the love dart mechanism is incorporated to update snail positions in each iteration. These concepts, along with their respective mathematical formulations, are further elaborated in the subsequent steps. The SHMS flowchart is presented in Fig. 3

### 3.1. Framework of SHMS algorithm

Consider the generalized constrained optimization problem in minimization sense as follows:

$$\min f(X) = f(X_1, \ldots, X_i, \ldots, X_N) \quad i = 1,2,3, \ldots, N \tag{9}$$

Subject to

$$g_i(X) < 0, i = 1,2, \ldots, n$$

$$h_j(X) = 0, j = 1,2, \ldots, m$$

$$\psi = X_i^l \leq X_i \leq X_i^u$$

Step 1: Generation of Homes and Snails

Initialize the position of $H$ homes $X_H = (X_1, \ldots, X_h, \ldots, X_H)$. Calculate the intervals $\psi_S$ around the position $X_H$ to generate the $S$ snails in the close neighbourhood of every home $h$:

$$\psi_S = X_H + c$$

where $c$ is sampling interval parameter which the keep the positions of the snail within the close neighbourhood of its home. Every snail randomly generates its position within the interval $\psi_S$ as follows:

$$X_S^h = \left((X_1), \ldots, (X_s), \ldots, (X_S)\right)^h$$

The snail solutions associated with every home $h, (h = 1, \ldots, H)$ are calculated as follows:

$$\left(f(X_1^h), \ldots, f(X_s^h), \ldots, f(X_S^h)\right)$$

Step 2: Apply the penalty function approach for constraint handling
The Static Penalty Function approach is used to handle the constraints. It is mathematically expressed as follows:

$$PF = \theta \times \left(\sum_{i=1}^{n} g_i(X) + \sum_{j=1}^{m} h_j(X)\right) \tag{10}$$

where, $\theta$ is penalty parameter. It required preliminary trials to set an appropriate value of $\theta$.
Further, the pseudo-function is calculated as follows:

$$F(X_s^h) = f(X_s^h) + PF \tag{11}$$

Step 3: Calculate the Fecundity Index

The fecundity index measures the snail's reproductive readiness, increases when its solution to a problem consistently improves. Conversely, it decreases when the solution deteriorates. A positive index signifies the snail is ready to mate, while a negative index indicates it is moving towards a worse solution and is not receptive to mating. It is measure as follows:

$$I_{S^h} = \left\|\frac{F(X_s^h)^{iter} - F(X_s^h)^{iter-1}}{F(X_s^h)^{iter} - F(X_s^h)^{iter-2}}\right\| \tag{12}$$

If $I_{S^h} \neq 0$ else $I_{S^h} = rand(0,1)$

Eq (10) is applicable in third iteration. For the first two iterations the $I_{S^h}$ is randomly generated between (0,1).

Step 4: Calculate the probability and apply the Roulette wheel approach for the distribution of snails

The probability of the snails is calculated as follows:

$$P_S = \frac{\frac{1}{F(X_s^h)}}{\sum_{s=1}^{S} \frac{1}{F(X_s^h)}} \tag{13}$$

After calculating the probability of each snail, the cumulative sum of those is calculated which defines the space occupied by the snails on the Roulette wheel. Snails with better function values take up more space, increasing their chances of selection. Further, a random value between 0 and 1 is generated for each snail to determine to which snail it will perform mating.

Step 5: Mating and Love Dart

The mating process in SHMS algorithm take place when snails use a sharp, calcareous structure that inject hormone into the fecund snail. That promotes successful storage of sperms of the love dart shooting snail into the fecund snail. The fecundity of the snail is denoted by $I_{S^h}$, which suggests how ready is the snail for reproduction. The mathematical expression of $LD$ stated below provides the step size for the next iteration. If the difference of the current and fecund snail fitness function is small, it will give very small step size, which will result in slow convergence. The snail solutions will move very slowly towards an optimum solution. In terms of snail biology, it is an analogy of unsuccessful storage of the donor snail's sperms stored in the recipient's sack. Higher is the value of $LD$, higher is the chances of sperm storage and hence successful mating and fertilization in the recipient's body.

$$LD_S = \frac{1}{I_{S^h} \times \left(F(X_S^h) - F\left(X_{S_{fecund}}^h\right)\right)} \qquad (14)$$

The large values of the love dart matrix are scaled down between zeroes and one. The normalization of the LD matrix is calculated as follows:

$$LD_{Normal} = \frac{LD_S - (LD_S)_{min}}{(LD_S)_{max} - (LD_S)_{min}} \qquad (15)$$

Step 6: Trail following and position update

The snail having better solution carries more weightage as compared with other snails. Generally, snails follow the trail/cues while returning to their home. There is a possibility that while returning to their home the snail may follow trail/ cues generated by other fecund snails $s_{fecund}^h$. It may happen that the snail $s^h$ can get the home of other snails. According to that, the snail generates a new position in the vicinity of the current home and continue to travel. The new position is identified as follows:

$$s_{up}^h = LD_{normal} \times (s^h - s_{fecund}^h) \qquad (16)$$

In this paper, the stepped power transmission shaft is proposed and solved using SHMS algorithm. The algorithm is coded in Google Colab with Python language and the simulations are run on Windows platform using 6$^{th}$ Gen Intel Core i5 6200U @ 2.3GHz processor speed and 8 GB RAM. The SHMS algorithm is run for 30 times and statistical results are noted and compared with the solution obtained ANSYS Workbench.

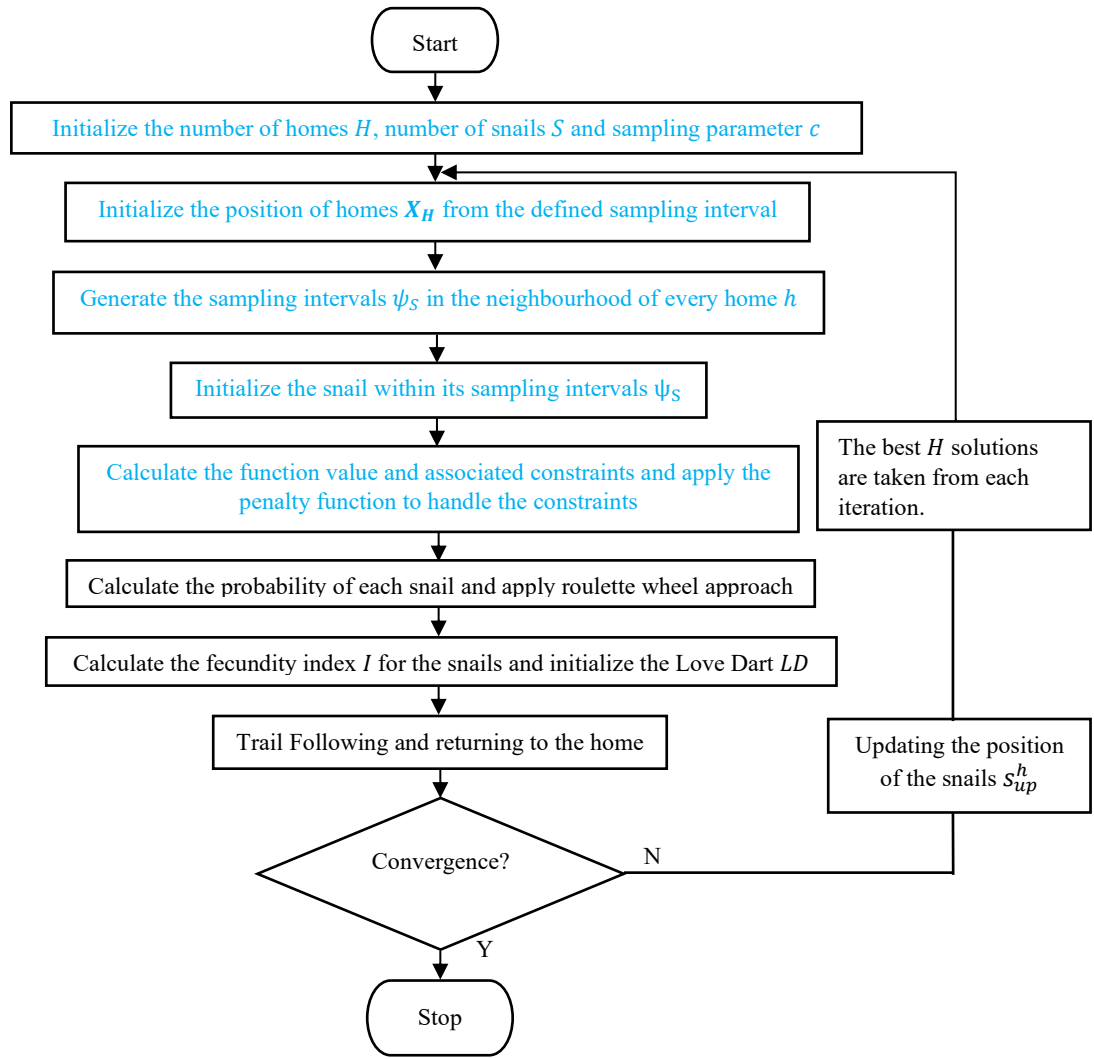

**Fig. 3** Flowchart of the SHMS Algorithm

## 4. Stepped transmission shaft problem

For the purpose of optimization of mechanical component, one must describe its physical behaviour based on certain mechanical principles. The process begins by defining a particular mechanical system and considering the required material properties, external forces boundary considerations and governing equations, such as Newton's second law, Hooke's Law, for Elasticity, Modified Goodman Criteria for combined loading, etc. Those governing equations needs to be solved in order to obtain the performance prediction, optimised weights and dimensions and simulate those data in Simulation software. In this paper, the design of stepped transmission shaft is proposed. The detailed mathematical formulation is presented in Section 3.1.

### 4.1 Problem Statement

The stepped transmission shaft, made of AISI 1040 CD Steel, density 0.2834 $lb/in^3$ consists of 3 stepped sections with lengths 5.9, 10 and 10 $inches$ respectively. The shaft is responsible for transmission of power from a motor to another system through meshing of gears. The transmission shaft has two rotating elements; a V-groove pulley of pitch diameter 20 $inches$ weighing 60 $lbf$ and a spur gear with pitch diameter of 10 $inches$ weighing 25 $lbf$. A motor of radius 6 $inches$ rotating at 1200 $RPM$ is belted to the V-groove pulley, the pulley being connected to the shaft by means of keys, which transmits power that fluctuates between 5 $hp$ to 20 $hp$. There are two pillow block bearings which provide support and provide restrictions in the lateral and longitudinal direction; one at extreme right end of the shaft and another at the left side between the gear and pulley, as shown in Fig. 4. The belt passes through the V-groove pulley with a V-groove angle 90°, and the coefficient of friction between the belt and the pulley is 0.31 to maintain the tension between the tight and slack side. The gear, on the other hand, has a pressure angle of 20°. The optimum diameter values are needed to calculate for each stepped section such that the deflection is minimized.

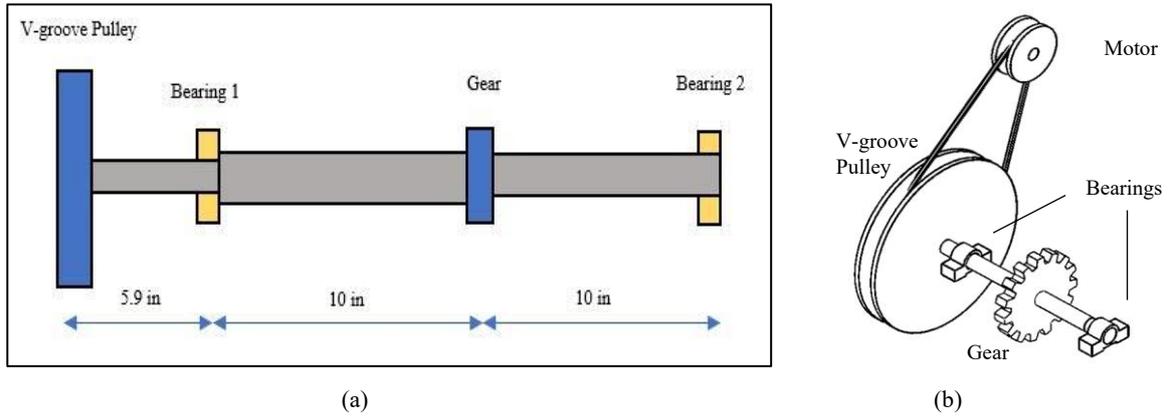

(a)                 (b)

**Fig 4** (a) Diagrammatic representation of the rotating elements in a stepped transmission shaft
(b) Actual Model design with its rotating elements and bearing supports

**4.2 Formulation of Stepped transmission shaft problem using SHMS**

Mathematical modelling of metaheuristic algorithms involves framing an optimization problem using objective functions and constraints, aiming to efficiently discover near-optimal solutions. The objective function $f(x)$ quantifies the performance measure to be optimized (either minimized or maximized) within a defined solution space. The stepped transmission shaft is designed to minimize the overall weight. The objective function is written as follows:

$$Min\ f(x) = W = \gamma \frac{\pi}{4}(L_1 d_1^2 + L_2 d_2^2 + L_3 d_3^2) \tag{17}$$

The value of $\gamma = 0.2834\ lb/in^3$, the lengths of the different sections are taken as $L_1 = 5.9\ in$ whereas $L_2 = L_3 = 10\ in$. The constraints $g_1, g_2, g_3$ gives the design diameter obtained by solving the Modified Goodman Equation for combined loading.

$$g_j = d_j - k(d_j) \leq 0 \quad for\ j = 1,2,3 \tag{18}$$

Difference between the adjacent diameters $d_2$ and $d_1$ must be more than 0.0787 (Rodriguez-Cabal et al., 2018).

$$g_4 = d_2 - d_1 \geq 0.0787 \tag{19}$$

Difference between adjacent diameters $d_2$ and $d_3$ must be more than 0.0787 (Rodriguez-Cabal et al., 2018).

$$g_5 = d_2 - d_3 \geq 0.0787 \tag{20}$$

Deflection of the shaft must less than $0.005\ in$ (Cabal et al., 2019).

$$g_6 = |y| \leq 0.005 \tag{21}$$

$$k(d_j) = \left[\frac{32 N_f}{\pi} \cdot \left\{\frac{\sqrt{(K_f M_a)^2 + \frac{3}{4}(K_{fs} T_a)^2}}{S_e} + \frac{\sqrt{(K_{fm} M_m)^2 + \frac{3}{4}(K_{fsm} T_m)^2}}{S_{ut}}\right\}\right]^{1/3} \tag{22}$$

$$y = \left\{ \frac{\frac{F_a \langle x \rangle^3}{6} + \frac{F_b (x-5.9)^3}{6} + \frac{F_c (x-15.9)^3}{6} + C_1 x + C_2}{EI} \right\} \quad (23)$$

The calculation of the deflection $y$ is carried out using Macaulay Method of slope and deflection which consist of a special type of function called Ramp function. The function results in zero if the value of $x$ is less or equal to $a$, and evaluated only when $x$ value is greater than $a$. The function is defined as follows:

$$\langle x - a_i \rangle = \begin{cases} 0 & if \ x < a_i \\ (x - a_i)^n & if \ x \geq a_i \end{cases} \quad (24)$$

$$for \ n \geq 0$$

The equation of moment is written as

$$EI \frac{d^2 y}{dx^2} = M = F.x$$

(23)

So, the deflection is calculated by double integrating the LHS and equating it to $y$ (Egelhoff and Odom, 2014, Rodriguez-Cabal et al., 2021, Rodriguez-Cabal, 2018, Guzmán and Delgado, 2005).

$$y = \int \left[ \int \frac{M}{EI} dx \right] dx + C_1.x + C_2 \quad (25)$$

$$y = \frac{\frac{F x^3}{6} + C_1 x + C_2}{EI} \quad (26)$$

The values of $C_1$ and $C_2$ are calculated using the boundary condition, where $y = 0$ at $x = 5.9$ and $25.9 \ inches$. The values of $K_{fsm}$ and $K_{fs}$ are taken as 2.05 and 2.05 respectively. The values of $K_f$ and $K_{fm}$ for the diameters $d_1$, $d_2$ and $d_3$ are 2.05, 3 and 2.4 respectively. These values are calculated using Norton Theory (Guzmán and Delgado, 2025). Table 1 provides the constants considered for calculating the diameters and deflection in the constraint equations. To addressed these constraints a well-known Static Penalty Function (SPF) (Kale and Kulkarni, 2018, 2021a) approach is used. This ensures that when the solution approach towards an infeasible region, the penalty function ($PF$) imposes on the function value, guiding the solution towards feasible outcome. The PF is calculated as $PF = \sum_{i=1}^{n} g_i \times \theta$. This yields a modified objective function/pseudo-objective function $F(x)$ expressed as:

$$F(x) = f(x) + PF \quad (27)$$

where $\theta$ is a penalty parameter that controls the influence of the penalty term. This parameter needs several preliminary trials to set an appropriate value. The $\theta$ value is set to 100000.

**Table 1** Constant considered for the study of rotating elements and the stepped shaft

| Parameter | Value |
|---|---|
| Mass of Pulley | $60 \ lbs$ |
| Mass of Gear | $25 \ lbs$ |
| Maximum Shaft Torque ($T_{max}$) | $3500 \ lb - in$ |
| Minimum Shaft Torque ($T_{min}$) | $875 \ lb - in$ |
| Shaft Pulley Radius | $20 \ in$ |
| Shaft Pulley rotation speed | $360 \ RPM \approx 38 \ rad/s$ |
| Shaft Pulley V-groove angle ($2\beta$) | $90°$ |
| Shaft pulley wrap angle ($\alpha$) | $210° = \frac{7\pi}{6} rad$ |
| Coefficient of friction between belt and pulley ($\mu$) | 0.31 |
| Pressure angle of Gear ($\phi$) | $20°$ |
| Gear Pitch diameter | $10 \ in$ |
| Modulus of Elasticity for AISI 1040 CD Steel | $29000000 \ psi$ |
| Density for AISI 1040 CD steel | $0.2834 \ lb/in^3$ |
| Safety Factor $N_f$ | 2.2 |
| Tensile Yield Strength ($S_y$) | 70300 $psi$ |
| Ultimate Tensile Strength ($S_{ut}$) | 75000 $psi$ |
| Fatigue Strength chosen ($S_e$) | 24314.3354 $d_i^{-0.097}$ $psi$ |

## 5. Analytical and FEA Analysis

The Shaft consists of rotating elements such as V-groove pulley and a spur gear of pitch diameter 20 $in$ and 10 $in$ respectively. The shaft rotates anticlockwise with a speed of 360 RPM, and the maximum torque transferred is 3500 $lb - in$. The gear produces a counter torque to overcome the forces from the meshed gear. The loads produced by the rotating components on the shaft are applied and Bending Moment diagrams and constructed using the Mechanical Software MDSOLIDS. The results of deflection corresponding to the diameters which obtained from SHMS algorithm are validated using ANSYS Workbench, a well-renowned FEA Tool which excels in static, dynamic and transient structural analysis of Mechanical components.

### 5.1 Analytical Approach

The values of alternating and mean Bending moments can be obtained from the vertical plane cases of the bending moment diagrams obtained by solving the static conditions.

The alternating and mean components of the Torque ($T_a$ and $T_m$ respectively) is calculated as:

Speed of Motor (driver) pulley $N_1 = 1200\ R.P.M.$

Maximum Power ($P_{max}$) = 20 hp = 132000 $lbf - in/s$

Minimum Power ($P_{min}$) = 5 hp = 33000 $lbf - in/s$

Maximum Torque ($T_{max}$) can be calculated using the Power equation

$$P = \frac{2\pi N T}{60} \tag{28}$$

Maximum Torque ($T_{max}$) = 1050.36 $lbf - in$

Minimum Torque ($T_{min}$) = 262.60 $lbf - in$

In the shaft, there will be fluctuations of loads due to fluctuating torque, so there will be mean and alternating torques developed as well, which can be calculated as

$$T_{mean} = \frac{T_{max}+T_{min}}{2} = \frac{3500+875}{2} = 2187.5\ lbf - in \tag{29}$$

$$T_{alt} = \frac{T_{max}-T_{min}}{2} = \frac{3500-875}{2} = 1312.5\ lbf - in \tag{30}$$

For calculating the forces coming from the pulley, we consider the ratio of tensions of the v-belt drives tight side ($T_1$) and slack sides ($T_2$) are given as

$$\frac{T_1}{T_2} = e^{\mu\alpha\ \cosec\beta} \tag{31}$$

Also, to calculate the torque transferred can be written as

$$T_1 - T_2 = \frac{Torque}{Radius} \tag{32}$$

Thus, using free body diagram the value of $T_1$ and $T_2$ can be determined, which needs to be further resolved due to the arrangement of the motor and Pulley. The $T_1$ and $T_2$ tensions for maximum and minimum case, after resolving will give the components along vertical and horizontal direction. The angle between the motor and the pulley comes out to be $\tan^{-1}\frac{25}{10} = 68.199°$. The maximum cases and minimum case for the tensions in tight and slack side are resolved as shown in Fig. 5 (a).

On the other hand, for gear load calculation, considering the Maximum Torque transfer case, torque transferred by the gear is $3500\ lbf - in$. The tangential force $F_T = \frac{3500}{5} = 700\ lb$ (Horizontal plane). The radial force is given by $F_R = F_T \tan 20° = 257.779\ lbf$ (Vertical plane).

In the Minimum Torque transfer case, torque transferred by the gear is $875\ lb - in$ as transferred by the pulley. The tangential force $F_T = \frac{875}{5} = 175\ lbf$ (Horizontal plane) and the radial force is given by $F_R = F_T \tan 20° = 63.695\ lb$ (Vertical plane). The resolved forces are presented in Fig. 5 (b). These values of forces in the vertical and horizontal direction for the maximum and minimum case are used to calculate the bending moment diagram using MDSOLIDS.

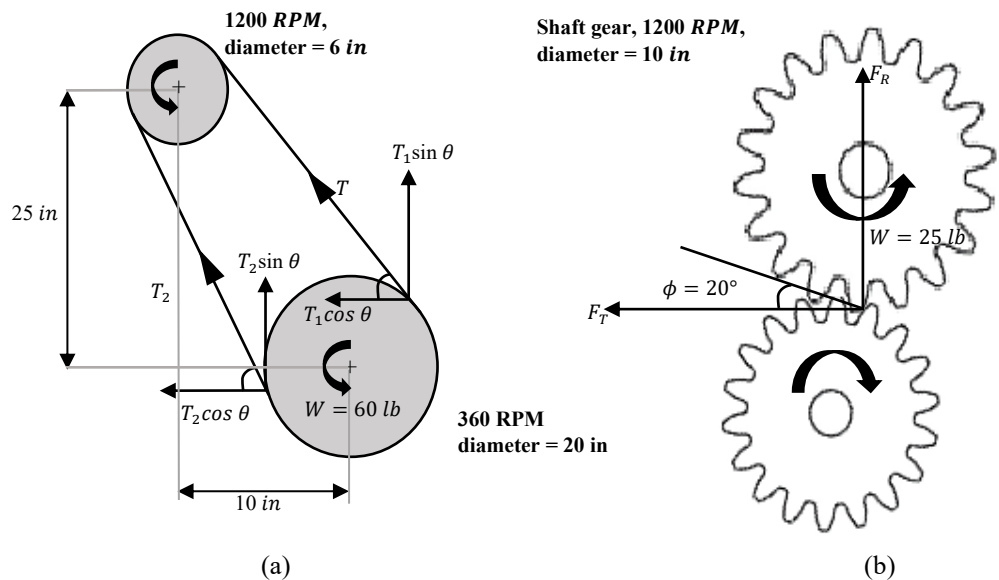

**Fig. 5** Free body diagram for the forces exerted by the rotating elements on the different sections of the stepped shafts
(a) The V-groove pulley generates a vertical and horizontal force from the tensions $T_1$ and $T_2$, considering the weight of pulley to be $60\ lbs$
(b) The spur gear generates counter forces due to the gear meshed with it, in vertical and horizontal plane, considering the weight of the gear to be $25\ lbs$

The vertical and horizontal forces are evaluated using free body diagram for both maximum and minimum cases (Fig 10(a) and Fig 10(b)) and the bending moment for the vertical and horizontal forces are calculated. The loads at the points of the V-groove pulley and the spur gear on the shaft are analytically calculated (refer Table 2). The detailed analytical procedure is presented in APPENDIX A. These loads are further used in MDSOLIDS (version 4.1.0) to obtain the Load Diagram and Bending Moment Diagram (BMD) at the points of both the rotating elements pulley and gear (refer Fig 6). The module of determinate beams is specifically used for this application. An overhanging beam of length $25.9\ in$ was designed using the simply supported beam with fixed supports at $5.9\ inches$ and $25.9\ in$, respectively.

To have a visual understanding of the direction of the forces, the line diagrams that represent the forces in vertical and horizontal planes are considered. For the free body diagrams, it is seen that for the vertical loads, the upwards forces are considered positive and downward forces are considered negative with respect to $y$ axis, as shown in Fig. 10(a). The values of $427.45\ lb$ and $229.78\ lb$ are the vertical forces in the upward direction and along the positive $y$ direction excluding the weight of the pulley ($60\ lb$) and gear ($25\ lb$), respectively. Therefore, the reaction force(s) are counter these forces in the opposite direction, hence taken as negative values. Similarly, in the horizontal plane, the values along positive $x$ axes are taken to be positive and negative when the forces are in the opposite of $x$ axis. Thus, the values of $194.98\ lb$ and $700.1\ lb$ are considered negative and its corresponding reactive force(s) will be negative. This is also considered for the minimum condition as shown in Fig. 10(b), and the values are presented in Table 2.

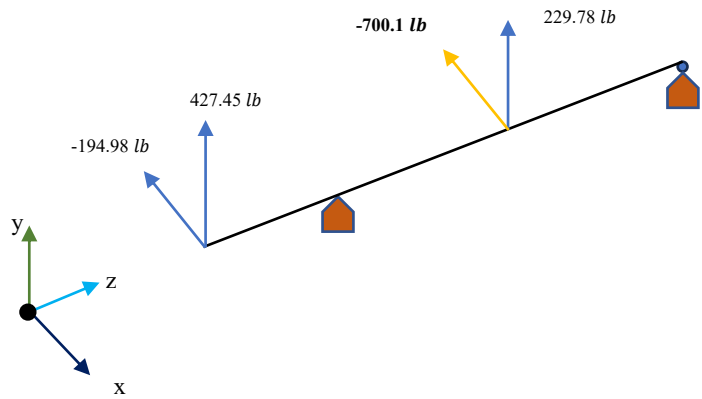

**Fig 10(a)** Free body diagram for the forces on the vertical and horizontal planes in the maximum power transfer condition

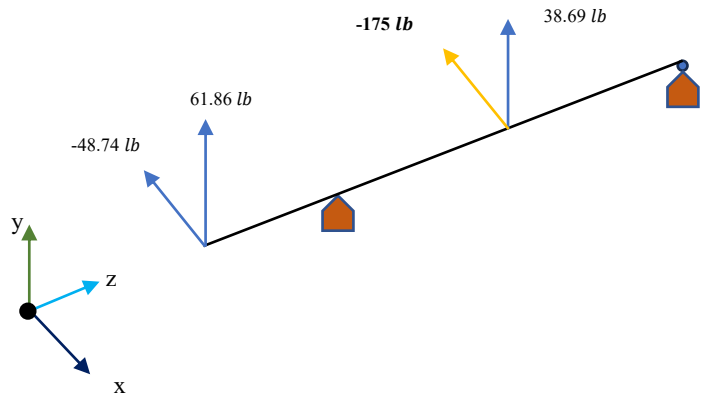

**Fig 10(b)** Free body diagram for the forces on the vertical and horizontal planes in the minimum power trasnfer condition

Table 2: Load acting of the shaft and corresponding reactions forces

| Condition | Plane | P1 (Pulley) | Reaction 1 at support A | P2 (Gear) | Reaction 2 at support B |
|---|---|---|---|---|---|
| Maximum | Vertical | 427.45 lb | -668.44 lb | 229.78 lb | 11.21 lb |
|  | Horizontal | -194.98 lb | 602.55 lb | -700.1 lb | 292.53 lb |
| Minimum | Vertical | 61.86 lb | -99.46 lb | 63.695 lb | -1.10 lb |
|  | Horizontal | -48.74 lb | 150.62 lb | -175 lb | 73.12 lb |

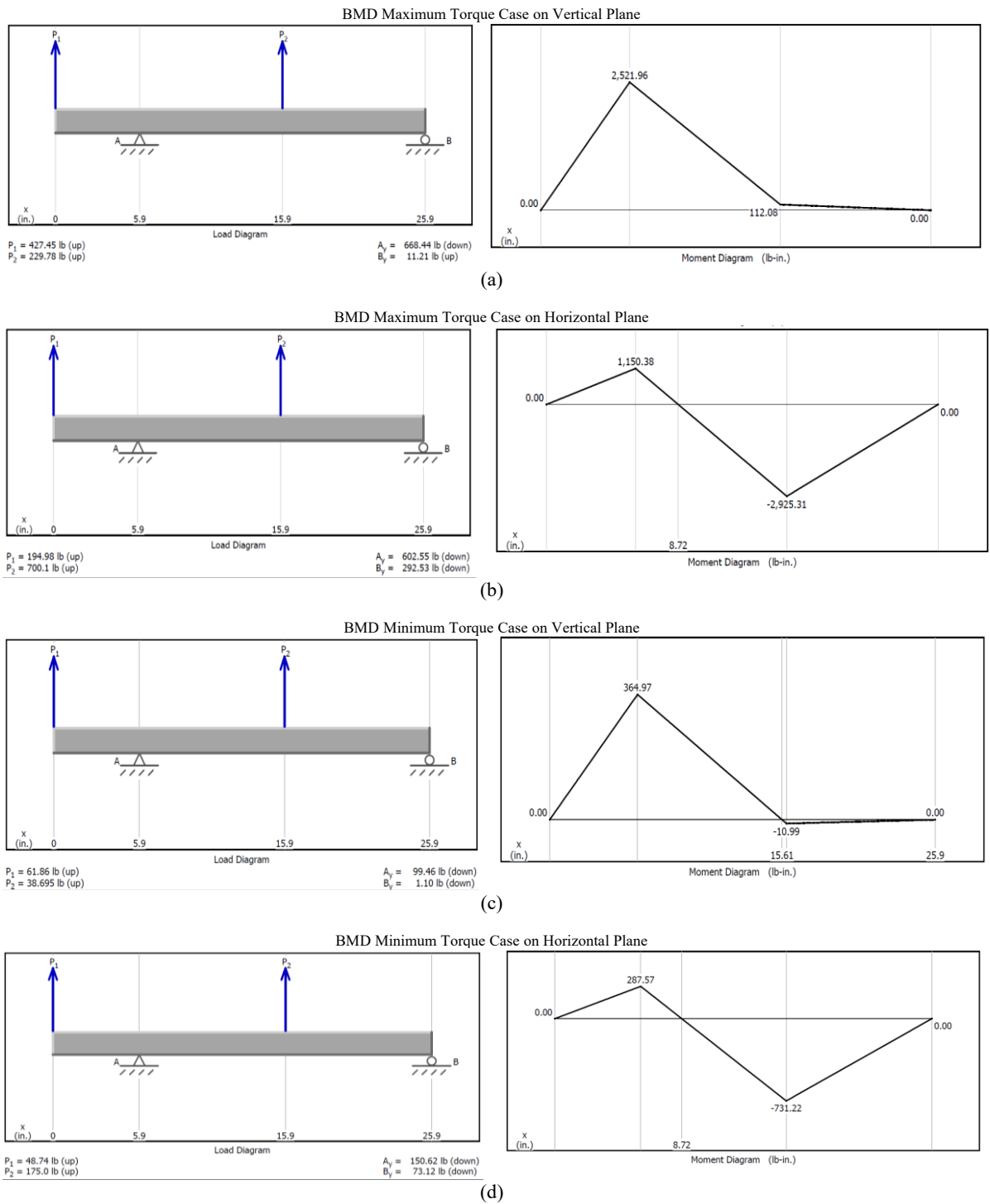

**Fig. 6** Force diagrams and Bending moment diagrams evaluated using MDSOLIDS software. Forces in pounds ($lb$) and bending moment in pound-inches ($lb-in$)
(a) Force and Bending Moment diagram for Maximum power transfer case in Vertical Plane
(b) Force and Bending Moment diagram for Maximum power transfer case in Horizontal Plane
(c) Force and Bending Moment diagram for Minimum power transfer case in Vertical Plane
(d) Force and Bending Moment diagram for Minimum power transfer case in Horizontal Plane

### 5.2 ANSYS Workbench Setup

From the analytical process the force and moment acting on the shaft are resolved using Load Diagram and BMD. Further, the ANSYS Workbench 2022 R1 is used for Finite Element Analysis (FEA) simulation with Static Structural solver (Rodriguez-Cabal, 2018). The diameters ($d_1, d_2, d_3$) obtained using SHMS algorithm are utilized in ANSYS SpaceClaim to create a 3D Model. AISI 1040 CD Steel is assigned as a material, with its properties referred from MatWeb materials database (MatWeb, 2020). The input moment of 3500 $lb-in$ in anticlockwise direction (maximum torque transfer condition) from the motor as well as the counter torque experienced in the clockwise direction has been taken into account. The rotational velocity of 360 $RPM$ (converted to 38 $rad/s$) has been applied to the whole body of the stepped shaft. The cylindrical supports are provided at positions 5.9 $inches$ and 25.9 $in$ of the transmission shaft (refer Fig. 7). The static conditions are also taken in horizontal and vertical directions as forces at the pulley and gear positions. Mesh size has been selected as 0.2 $in$ with default element type of tetrahedral.

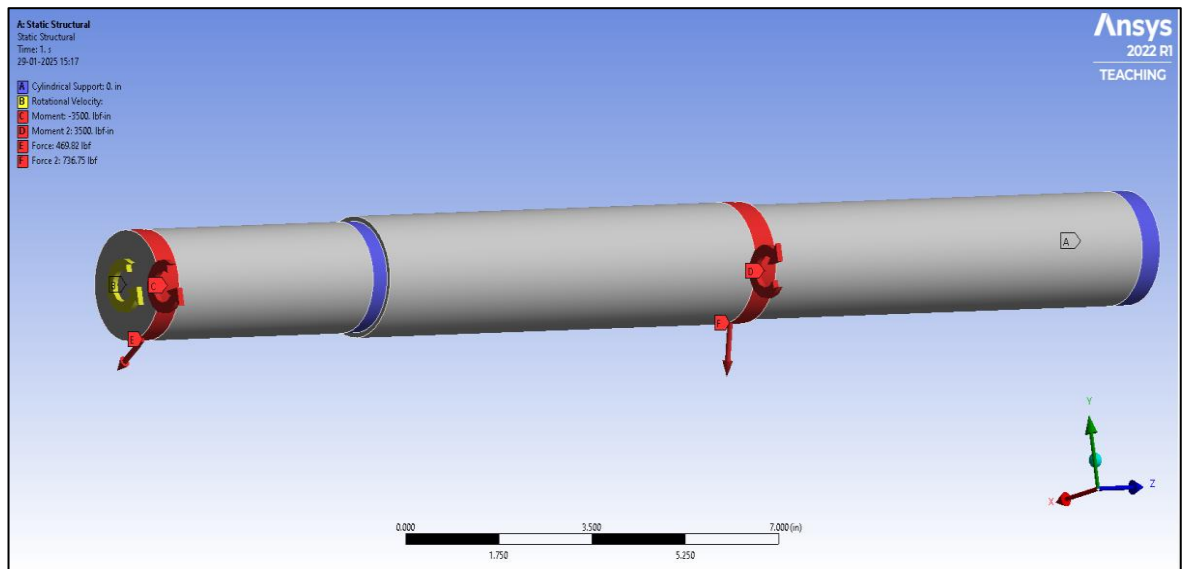

**Fig. 7** Loads and Boundary conditions applied on the stepped shaft of material AISI 1040 CD Steel

### 6. Results Discussion

The proposed stepped transmission shaft aims to minimize the overall weight is successfully solved using the SHMS algorithm. The constraint associated are handled with the SPF approach. The SHMS algorithm is run for 30 times to obtain the statistical results. From that the best, mean and worst values of overall weight ($W$) and the associated diameters of stepped transmission are presented in Table 2. The average function evaluations, standard deviation and average CPU time reported are 937, 0.7062 and 2.11 $sec$, respectively. The statistical results clearly demonstrate the robustness of the SHMS algorithm. The convergence plot for the best run is presented in Fig. 8. The results achieved using SHMS algorithm is very promising as it gives the optimum diameters and the minimum possible weight of the stepped shaft. These results can be considered as a benchmark for this problem.

**Table 2** Statistical results obtained from SHMS algorithm showing the best, mean and worst weights with its corresponding diameters

| Result | $W$ ($lbf$) | $d_1$ ($in$) | $d_2$ ($in$) | $d_3$ ($in$) |
|---|---|---|---|---|
| Best | 16.5139 | 1.6822 | 1.8105 | 1.5721 |
| Worst | 19.6995 | 1.7654 | 1.9289 | 1.8139 |
| Mean | 17.0795 | 1.7086 | 1.8291 | 1.6118 |

The corresponding diameters $d_1, d_2$ and $d_3$ of best, worst and mean weight of the shaft obtained from SHMS algorithm are further used to create a CAD model in ANSYS SpaceClaim and simulate it with the obtained loads and boundary conditions on ANSYS Workbench. Further, the structural analysis of the stepped transmission shaft is conducted on ANSYS Workbench to identify the maximum deflection. The analysis plots presented in Fig 9(a), Fig 9(b) and Fig 9(c) illustrate the maximum deflection corresponding to the best, worst and mean results obtained of SHMS algorithm. It must be noted that at the distance of 15.9 $in$ the maximum deflection needs to be reported where the maximum bending moment occurs (Fig 6). It is successfully achieved in the two cases best and mean. However, the maximum deflection in worst case is appeared at the start point of the shaft.

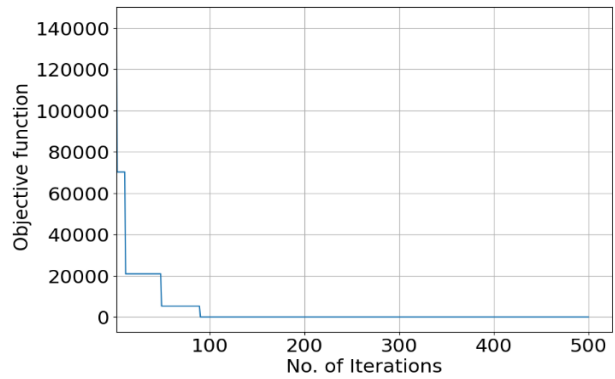

**Fig. 8** The SHMS Algorithm convergence towards the optimum best solution (16.5139 *lb*)

The diameters obtained from the SHMS algorithm produces satisfactory results with optimum diameters for each stepped section, and thus giving the corresponding weights. The shaft, with necessary loads and boundary conditions produces deflection 0.0044 *in* less than the allowable deflection of 0.005 *in*, as required by the design criteria, at and around the gear mounted point. The following ANSYS Workbench plots provides the visual verification of SHMS Algorithm results that minimises the deflection at the gear point below 0.005 inches, by applying the necessary loads, moments and boundary conditions.

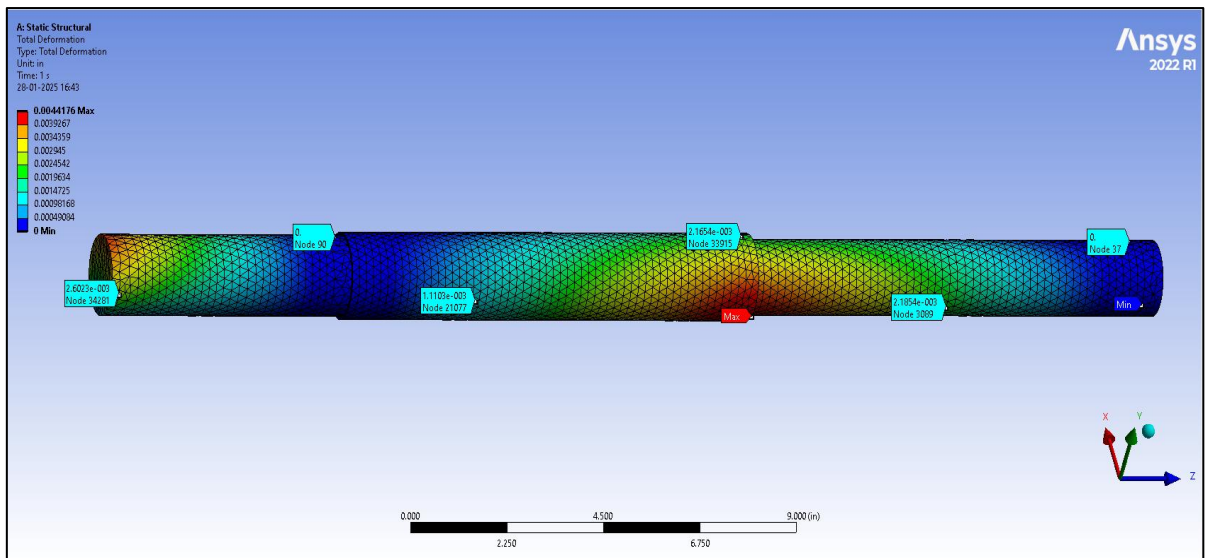

(a)

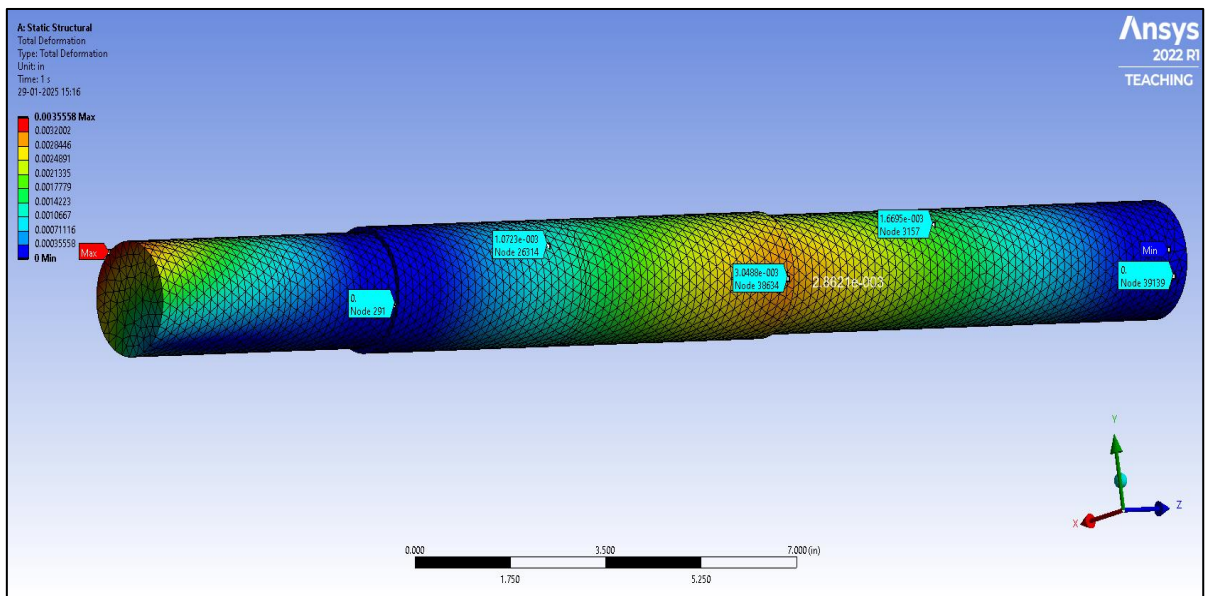

(b)

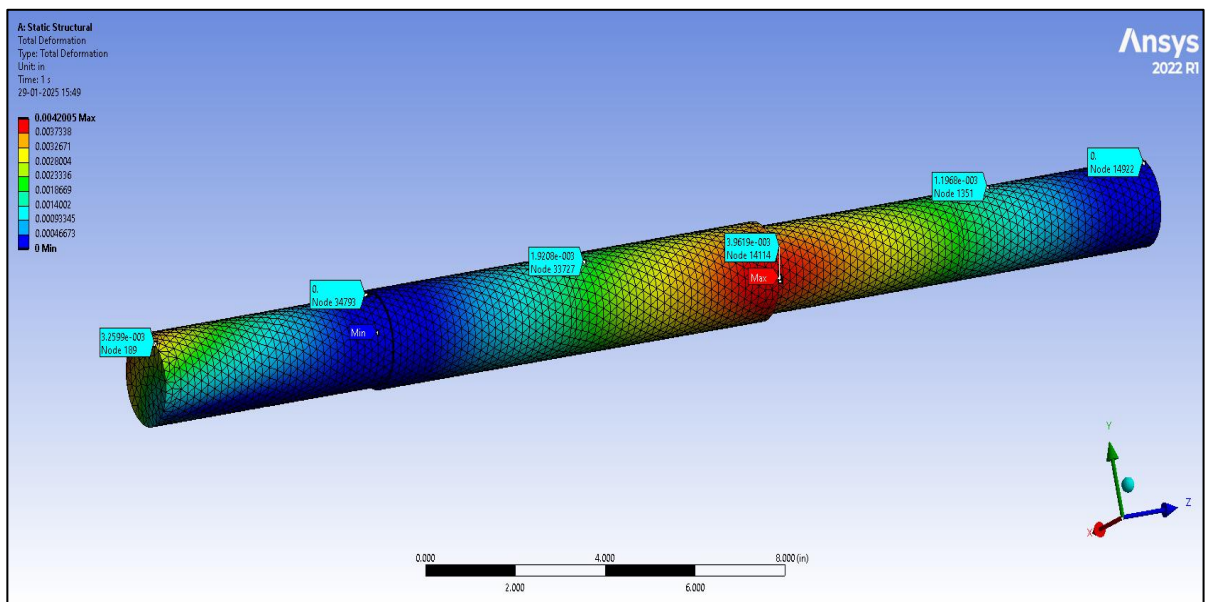

(c)

**Fig. 9** ANSYS Workbench results for the deflections of the stepped shaft at the Gear mounted location with three results shown by the contour plots
(a) Best Solution, $d_1$=1.6822 $in$, $d_2$=1.8105 $in$, $d_3$= 1.5721 $in$ and Weight = 16.5139 $lb$
(b) Worst Solution, $d_1$=1.7654 $in$, $d_2$=1.9289 $in$, $d_3$=1.8139 $in$ and Weight =19.6995 $lb$
(c) Mean Solution, $d_1$=1.7086 $in$, $d_2$=1.8291 $in$, $d_3$= 1.6118 $in$ and Weight = 17.0795 $lb$

During the calculation of the deflection, the point where the gear is keyed to the shaft was considered at 15.9 inches, as the deflection at the gear point needed to be minimised such that the gear doesn't move away from its driven gear and result in inappropriate meshing and hence power transmission loss. This might also cause wear of gear teeth as the deflection would result in inappropriate contact. Hence, the deflection of shaft at the gear point is crucial for minimization. Table 3 shows a comparison between the deflection resulted by solving the SHMS algorithm and ANSYS Workbench. The SHMS algorithm considered only the deflection at the Horizontal plane since the priority of the algorithm was to reduce the bending moment (in turn, the deflection) wherever it is maximum. At 15.9 $in$, it is observed that the bending moment value comes out to be 2925.31 $lb - in$ due to the force of 700 $lb$. However, in ANSYS Workbench solution, both vertical and horizontal components of forces are taken into account, as the ANSYS Workbench uses the resultant of the horizontal and vertical components, giving accurate values of deflection. In both cases, it is observed that the worst case weight of the shaft produces the least deflection with 42% persent of error where best and the mean cases weight produces greater deflection with least error of 9% and 6% respectively. The errors in comparing the deflection occor due to the assumption of forces in the horizontal plane only, which in reality is the resultant of both vertical and horizontal forces, as well as fluctuation between maximum and minimum values of power transfer. A greater value of deflection will occur when it is experiencing the greater value of force. It is observed that on the Horizontal plane at maximum power transfer case, the value of the force of 700 $lb$ is generated which will cause deflection at the gear point. So, while evaluating using SHMS, the deflection value is greater than what is achieved using ANSYS Workbench. ANSYS Workbench does not have any calculation constraint, and both vertical and horizontal forces can be set up to achieve more accurate deflections.

Table 3 Comaprison of deflection values achieved from SHMS Algorithm and ANSYS Workbench

| Result | Deflection at 15.9 $in$ using SHMS | Deflection at 15.9 $in$ using ANSYS Workbench | Error (w.r.t. ANSYS) |
|---|---|---|---|
| Best | 0.00482 $in$ | 0.00442 $in$ | ≈ 9% |
| Worst | 0.00434 $in$ | 0.00304 $in$ | ≈ 42 % |
| Mean | 0.00446 $in$ | 0.0042 $in$ | ≈ 6% |

## 7. Conclusion

In this paper, the transmission stepped shaft problem is proposed and successfully solved using a metaheuristic algorithm known as SHMS algorithm and analyse using ANSYS Workbench. SHMS algorithm is a nature-inspired optimization algorithm mimics the biological behaviour of snails which continuously search for food, better shelter and the mates. This is possible by following the mucus trail available around. Snails follow trail of the fecund snail to perform mating and homing behaviour. Thus, the constant strives for survivability and sustainability of snails searching for better habitat and safety is what inspired the SHMS algorithm to gradually converge to the optimum value. The constraints associated with this problem are handled using a well-known static penalty function approach.

Summarizing the above studies, the proposed problem is considering the Modified Goodman usingVon Mises Stresses for fatigue analysis. This criterion is used to calculate the diameters of each section of the stepped shaft under various loading scenarios. The deflection is analysed using the Macaulay method, where ramp functions are used for slope and deflection calculation. If excessive deflection is allowed (i.e. > 0.005 $in$), the gear will misalign with its mating gear leading to bad contact and power loss during transmission. Hence, it is important to reduce the shaft deflection. In other words, to achieve the following, it is necessary to minimize both the weight of the shaft, and also the deflection of the shaft at the gear location. Furthermore, the force and bending moment calculation is conducted in MDSOLIDS. The statistical results such as best, mean and worst function value and associated diameters obtained form 30 trials of SHMS algorithm are utilized to create a CAD model using ANSYS SpaceClaim and analysis is carried out on ANSYS Workbench.

With the achieved values of mean and alternate bending moments, SHMS Algorithm gave optimum results as expected by reducing the deflection of the shaft at the gear region by providing the optimum values of diameters, as well as provided the minimum value of the weight of the shaft possible to satisfy all the constraint equations. Among the best, mean, and worst results, ANSYS Workbench verified the outcomes and provided the shaft model using AISI 1040 CD Steel. In Fig 10 illustrates, for the best case, the deflection was found to be 0.0042 $in$ with a corresponding weight of 16.5139 $lb$, which is less than the acceptable deflection limit of 0.005 $in$. Although it is considered the best solution in terms of optimization, the deflection value is greater than that of the worst-case scenario. In the worst case, the deflection is 0.00355 $in$, and the corresponding weight of the stepped shaft is 19.6995 $lb$. The larger diameter causes the problem of more self-weight, but due to more mass moment of inertia the deflection value gets reduced.

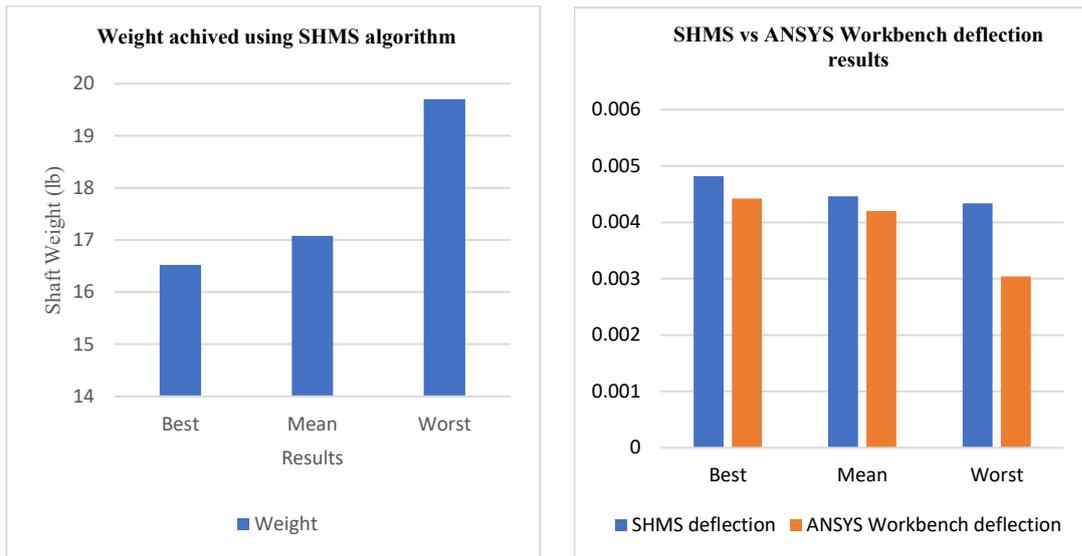

**Fig. 10** Comparison charts for demostrating different cases (Best, Mean and Worst)
(a) Weights achieved using SHMS algorithm for 3 cases
(b) Comparison of deflection in 3 cases for SHMS Algorithm and ANSYS Workbench

**Declarations**

- **Availability of data and material:** All the data and material are available in the manuscript.

- **Competing interests:** Authors have no competing and conflicting interests of any kind.

- **Ethical approval:** This article does not contain any studies with human participants or animals performed by any of the authors.

- **Funding:** No funding

- **Authors' contributions:**

  **Kaustav Shah:** Problem Formulation, Algorithm modification and implementation, Software analysis, Manuscript writing

  **Ishaan R Kale:** Problem Formulation, Algorithm Modification, Manuscript Writing, Manuscript Review

  **Vivek Patel:** Software analysis

  **Anand J Kulkarni:** Algorithm Modification, Manuscript Review

  **Puskaraj D Sonawwanay:** Supervision

**APPENDIX A**

**Resolution of Forces for Pulleys**

To resolve the forces of pulley, the ratio of tight and slack side can be written as

$$\frac{T_1}{T_2} = e^{\mu\alpha \operatorname{cosec} \beta}$$

Also, the relation between T1 and T2 can be written as

$$T_1 - T_2 = \frac{Torque}{Radius}$$

To obtain of T1 and T2 using the above equations

$$\frac{T_1}{T_2} = e^{\mu\alpha \operatorname{cosec} \beta}$$

<u>**Maximum case**</u>

Putting values of $\mu = 0.31, \alpha = 210° = \frac{7\pi}{6}, 2\beta = 90°$, we get

$$\frac{T_1}{T_2} = 4.987 \approx 5$$

$$\text{And } T_1 - T_2 = \frac{3500}{10} = 350$$

So, for maximum case T1 tight side tension and T2 slack side tension comes out to be

$$T_1 = 437.5 \ lb \ (up)$$
$$T_2 = 87.5 \ lb \ (up)$$

Resolving the vertical of both T1 and T2,

$$T_{1v} = 406.209 \ lb \ (up)$$
$$T_{2v} = 81.24 \ lb \ (up)$$

$$W_{pulley} = 60 \ lb \ (down)$$

Similarly resolving the Horizontal forces of T1 and T2

$$T_{1h} = 162.48 \ lb \ (in)$$
$$T_{2h} = 32.496 \ lb \ (in)$$

**<u>Minimum case</u>**

Putting values of $\mu = 0.31, \alpha = 210° = \frac{7\pi}{6}, 2\beta = 90°$, we get

$$\frac{T_1}{T_2} = 4.987 \approx 5$$

$$\text{And } T_1 - T_2 = \frac{875}{10} = 87.5$$

So, for minimum case T1 tight side tension and T2 slack side tension comes out to be

$$T_1 = 109.375 \ lb \ (up)$$
$$T_2 = 21.875 \ lb \ (up)$$

Resolving the vertical of both T1 and T2,

$$T_{1v} = 101.55 \ lb \ (up)$$
$$T_{2v} = 20.31 \ lb \ (up)$$

$$W_{pulley} = 60 \ lb \ (down)$$

Similarly resolving the Horizontal forces of T1 and T2

$$T_{1h} = 40.62 \ lb \ (in)$$
$$T_{2h} = 8.124 \ lb \ (in)$$